\documentclass[10pt,twocolumn,letterpaper]{article}

\usepackage[pagenumbers]{cvpr} 

\definecolor{cvprblue}{rgb}{0.21,0.49,0.74}
\usepackage[pagebackref,breaklinks,colorlinks,allcolors=cvprblue]{hyperref}
\usepackage{multirow} 
\usepackage[table]{xcolor}
\def\confName{CVPR}
\def\confYear{2026}

\title{Draft and Refine with Visual Experts}

\author{
    \begin{tabular}{cccc}
        SUNGHEON JEONG$^1$ & RYOZO MASUKAWA$^{1}$ & JIHONG PARK$^3$ & SANGGEON YUN$^1$
        \end{tabular}
        \\[0.7em]
        \begin{tabular}{cccc}
        WENJUN HUANG$^1$ & HANNING CHEN$^1$ & MAHDI IMANI$^2$ & MOHSEN IMANI$^1$
        \end{tabular}
        \\[1.0em]
        \begin{tabular}{c}
        $^{1}$University of California, Irvine, $^{2}$Northeastern University, $^3$MOLOCO \\
        {\tt\small sungheoj@uci.edu}
    \end{tabular}
}

\begin{document}
    \maketitle
    \begin{abstract} 
    While recent Large Vision–Language Models (LVLMs) exhibit impressive multimodal reasoning abilities, they often produce ungrounded, hallucinated responses by over-relying on linguistic priors rather than visual evidence. This critical limitation arises from the lack of a quantitative measure of how much these models actually rely on visual inputs during reasoning. We propose \textbf{Draft and Refine (DnR)}, an agent framework driven by a novel question-conditioned utilization metric. This metric quantifies the model’s actual reliance on visual evidence by first constructing a query-conditioned relevance map to localize question-specific evidence, and then assessing dependence through relevance-based probabilistic masking. Guided by this metric, the DnR agent refines its initial `draft' through targeted feedback from external visual experts. Each expert’s output (e.g., boxes, masks) is rendered as visual cues on the image, and the LVLM is re-queried to select the response that yields the greatest improvement in utilization. This process strengthens visual grounding of predictions without retraining or architectural changes. Experiments across a broad range of VQA and captioning benchmarks demonstrate consistent accuracy gains and reduced hallucination. These results show that quantifying visual utilization provides a principled path for designing more interpretable and evidence-driven multimodal agent systems that effectively leverage visual experts. Code is available at \mbox{\href{https://github.com/EavnJeong/Draft-and-Refine-with-Visual-Experts}{\texttt{Github}}}.
\end{abstract}
    
    \section{Introduction}
    \label{sec:introduction}
    Large vision language models (LVLMs) have begun to interface with diverse visual tools and experts, enabling complex reasoning that combines perception and language~\cite{shen2023hugginggpt, yang2023mmreact, wu2023visualprogramming, gupta2023viper, yao2023react, li2024visionllmv2, ye2024mplugowl2}. However, existing methods mainly rely on language-driven control, prompting models to call experts based on chain-of-thought or textual confidence~\cite{yao2023react, schick2023toolformer, shen2023hugginggpt, yang2023mmreact}. Such mechanisms inherit the biases and unreliability of the language model itself and seldom account for how effectively visual information is actually utilized~\cite{bai2024hallucination, favero2024multimodal}. Learning-based coordination frameworks, on the other hand, require costly and inflexible joint optimization across multiple experts and tasks~\cite{li2024mova, chen2024vcoder, li2024hyperllava, li2024sphinx}. This raises a fundamental question: \textit{can a vision–language model autonomously determine when and which visual expert to invoke—guided by its own perceptual needs rather than linguistic biases?}

    To address this challenge, we reconsider expert coordination from the perspective of visual utilization. Our initial intuition is simple: LVLMs should be guided to more effectively utilize visual information when forming its predictions \cite{li2023llava, yin2023mmhalbench}. However, not all visual cues are equally useful, because different tasks and questions require attention to different parts of the image \cite{liu2023groundingdino, kirillov2023segment}. Therefore, the key is not to maximize visual dependence globally but to evaluate how well the model utilizes the regions relevant to the given situation \cite{radford2021clip, chefer2021transformer}. Once the model identifies which visual information is most critical, the selected expert can provide additional evidence to refine the model's reasoning and improve its decision. Building on this idea, we develop a Draft and Refine (\textit{DnR}) framework that measures visual utilization and uses it as a quantitative criterion for selecting and incorporating visual experts. This formulation enables adaptive expert choice grounded in the model's actual perceptual behavior rather than relying on linguistic priors or heavy joint supervision.

    Existing LVLMs can describe what they see but cannot determine which visual elements are truly important or why they matter for a given task~\cite{selvaraju2019taking, xiao2024towards}. Because there is no explicit label or metric to assess such dependence, we first define a measurable criterion that quantifies how much the model relies on visual information through a relevance-based perturbation process. This enables us to evaluate visual utilization without additional supervision. After computing this utilization, we incorporate external visual experts to complement perceptual gaps by providing additional evidence. To align their outputs with the model’s perception, we introduce a visual rendering mechanism that highlights essential regions while suppressing irrelevant content. This controllable rendering mechanism enables flexible integration of diverse visual experts into general-purpose multimodal reasoning systems without retraining.

    Through this formulation, our approach offers a practical framework to quantify and leverage visual information within multimodal reasoning. It allows models to evaluate how effectively they use visual evidence and refine reasoning based on perceptual relevance rather than linguistic confidence. The rendering mechanism serves as a flexible interface for integrating diverse visual experts without retraining, enabling agent-style coordination that generalizes across domains. Although rendering styles and parameters require dataset- and model-specific tuning, the framework consistently improves performance across VQA and captioning benchmarks, showing a strong correlation with task accuracy and a substantial reduction in hallucination. Collectively, these findings suggest that measurable visual grounding offers a principled basis for developing interpretable and evidence-driven AI agent systems that leverage visual tools.

    \section{Related Work}
    \label{sec:related_work}
    
    \textbf{Multimodal Large Language Models.}
        Recent advances in multimodal large language models (MLLMs) have significantly expanded the capacity of language models to reason over visual inputs~\cite{li2023blip2, dai2023instructblip, liu2024llava16, huang2024kosmos2, alayrac2022flamingo, openai2023gpt4v, google2024gemini, li2025surveyvlm, yin2024surveymlmm}. By coupling pretrained vision encoders with powerful language backbones, these systems achieve strong zero-shot performance on captioning, visual question answering, and general visual reasoning tasks~\cite{villa2023merlim, li2025surveyvlm}. However, their reasoning process remains predominantly language-driven, relying heavily on internal linguistic priors rather than on visual grounding~\cite{favero2024multimodal, bai2024hallucination, villa2023merlim}. When the visual representation is coarse or misaligned, such models tend to hallucinate plausible yet unsupported content~\cite{bai2024hallucination, favero2024multimodal, villa2023merlim}. This imbalance between linguistic reasoning and perceptual understanding has been repeatedly observed across recent evaluations, which report limited grounding fidelity and reduced robustness on fine-grained or localized visual reasoning benchmarks~\cite{villa2023merlim, bai2024hallucination, li2025surveyvlm}. These findings underscore the need for frameworks that can explicitly assess, verify, and refine visual evidence throughout the reasoning process~\cite{bai2024hallucination, favero2024multimodal, li2025surveyvlm}.

    \noindent\textbf{Tool-Augmented and Agentic LLMs.}
        Recent research has transformed large language models into agentic systems that autonomously plan, execute, and integrate external tools or APIs for reasoning and perception~\cite{schick2023toolformer, yao2023react, shen2023hugginggpt, yang2023mmreact, liang2024octotools}. These approaches span programmatic reasoning agents that compose visual functions via code execution~\cite{gupta2023viper, wu2023visualprogramming, wu2023visualchatgpt}, tool-calling coordinators that orchestrate multiple pretrained experts in a modular pipeline~\cite{shen2023hugginggpt, yang2023mmreact, li2024metaprompting, li2024visionllmv2, ye2024mplugowl2}, and multimodal planners that integrate vision encoders for grounded decision-making~\cite{alayrac2022flamingo, li2023blip2, dai2023instructblip, liu2024llava16, huang2024kosmos2, openai2023gpt4v, google2024gemini, wang2024argus}. Despite their success, these systems generally rely on predefined or heuristic tool invocation, often selecting experts based on the LLM’s textual outputs or embedding similarities rather than measurable visual evidence~\cite{schick2023toolformer, shen2023hugginggpt, yang2023mmreact, li2024metaprompting, li2024visionllmv2, ye2024mplugowl2, li2024mova, chen2024vcoder, li2024hyperclip, li2024hyperllava, li2024sphinx}, and thus lack quantitative criteria for determining when visual assistance is necessary or which expert should be called. In contrast, our framework formulates expert invocation as a measurable decision process, guided by the model’s actual utilization of visual evidence, thereby enabling an adaptive and systematic bridge between perception and reasoning.
    
    \noindent\textbf{Hallucination Mitigation and Visual Grounding.}
        Addressing hallucination and weak grounding has become a central challenge in both language and vision–language modeling~\cite{bai2024hallucination,favero2024multimodal,li2025surveyvlm}. Existing approaches attempt to improve factual or visual alignment through faithfulness-oriented supervision~\cite{li2022faithfulvqa,han2023groundingattention}, retrieval-augmented reasoning~\cite{yang2023retrievalvqa,chen2024groundingqa}, or uncertainty-driven selection and abstention strategies~\cite{gurrolakim2022reliablevqa,khan2024selectvqa,kim2021activevqa,tuli2024semanticentropy}. While these methods reduce hallucination, they generally operate as post-hoc corrections or apply global feature amplification without explicitly conditioning on the query, often leading to over-attention on irrelevant regions and diluted reasoning fidelity~\cite{cqvqa2020,dontassume2021}. In contrast, our framework introduces a query-conditioned relevance map that directly links visual importance to question semantics, enabling selective and evidence-grounded refinement rather than uniform enhancement.
    \section{Draft and Refine with Visual Experts}
    \begin{figure*}
        \centering
        \includegraphics[width=1.0\linewidth]{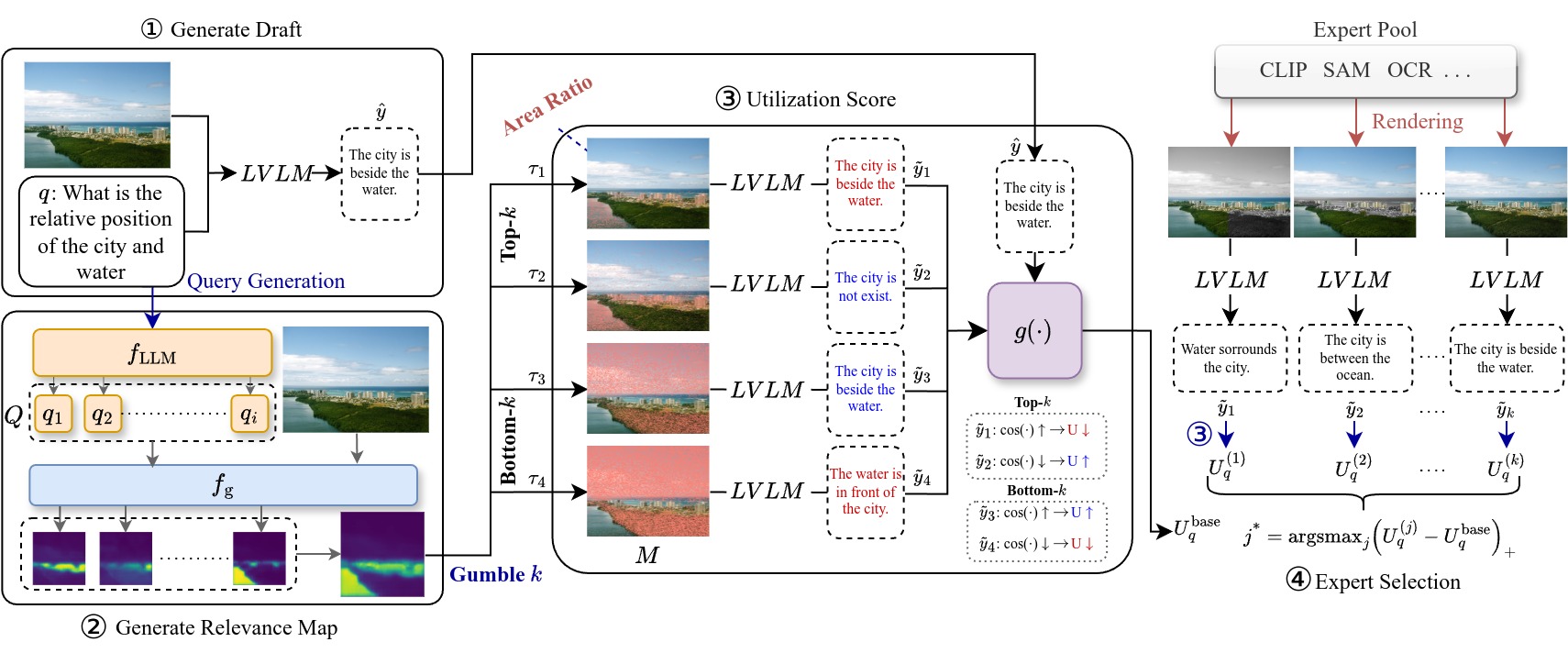}
        \caption{
            \textbf{Overview of the Draft-and-Refine (DnR) framework.} Given an image \(x\) and a question \(q\), the LVLM first generates an initial draft answer \(\hat{y}\) \textcircled{1}. The question is decomposed by \(f_{\mathrm{LLM}}\) into a query set \(Q=\{q_i\}\), and each query is grounded by \(f_g\) to produce spatial relevance maps, aggregated into \(r(x\mid q)\) \textcircled{2}. Gumbel-$k$ sampling masks Top-$k$ and Bottom-$k$ regions for perturbation, and a semantic encoder \(g(\cdot)\) measures similarity shifts between \(\hat{y}\) and perturbed predictions \(\tilde{y}_\tau\) to compute the utilization score \(U_q^{\mathrm{base}}\) \textcircled{3}. Expert models (e.g., CLIP, SAM, OCR) render structured visual evidence onto the image, producing refined outputs with updated utilization \(U_q^{(j)}\). The expert with the largest gain \(U_q^{(j)} - U_q^{\mathrm{base}}\) is selected for refinement \textcircled{4}.
        }
        \label{fig:dnr}
    \end{figure*}

    The Draft-and-Refine (\textit{DnR}) framework improves a LVLM by refining its draft response using specialized visual experts. Given an image $x$ and a question $q$, the LVLM first drafts an initial response based on its existing visual understanding. \textit{DnR} then analyzes the $x$ to identify regions most relevant to $q$, evaluates how the model’s reasoning depends on them, and selectively engages the expert that provides complementary visual evidence. By integrating this evidence, \textit{DnR} turns the LVLM’s passive description into an active, evidence-anchored reasoning process that steers its answers toward visually grounded decisions rather than incidental linguistic context.
    
    \subsection{Query-Conditioned Relevance Map}
        To assess how effectively the LVLM grounds its reasoning in visual evidence, we focus on identifying the regions of image $x$ that truly matter for a given question $q$. Rather than assuming that all parts of the image contribute equally, we recognize that only certain regions serve as meaningful evidence. Therefore, we introduce a \textit{query-conditioned relevance map} $r(x\mid q)$, which localizes and highlights the areas of the $x$ most informative for answering $q$.
        
\begin{figure}[t]
    \centering
    \includegraphics[width=1.0\linewidth]{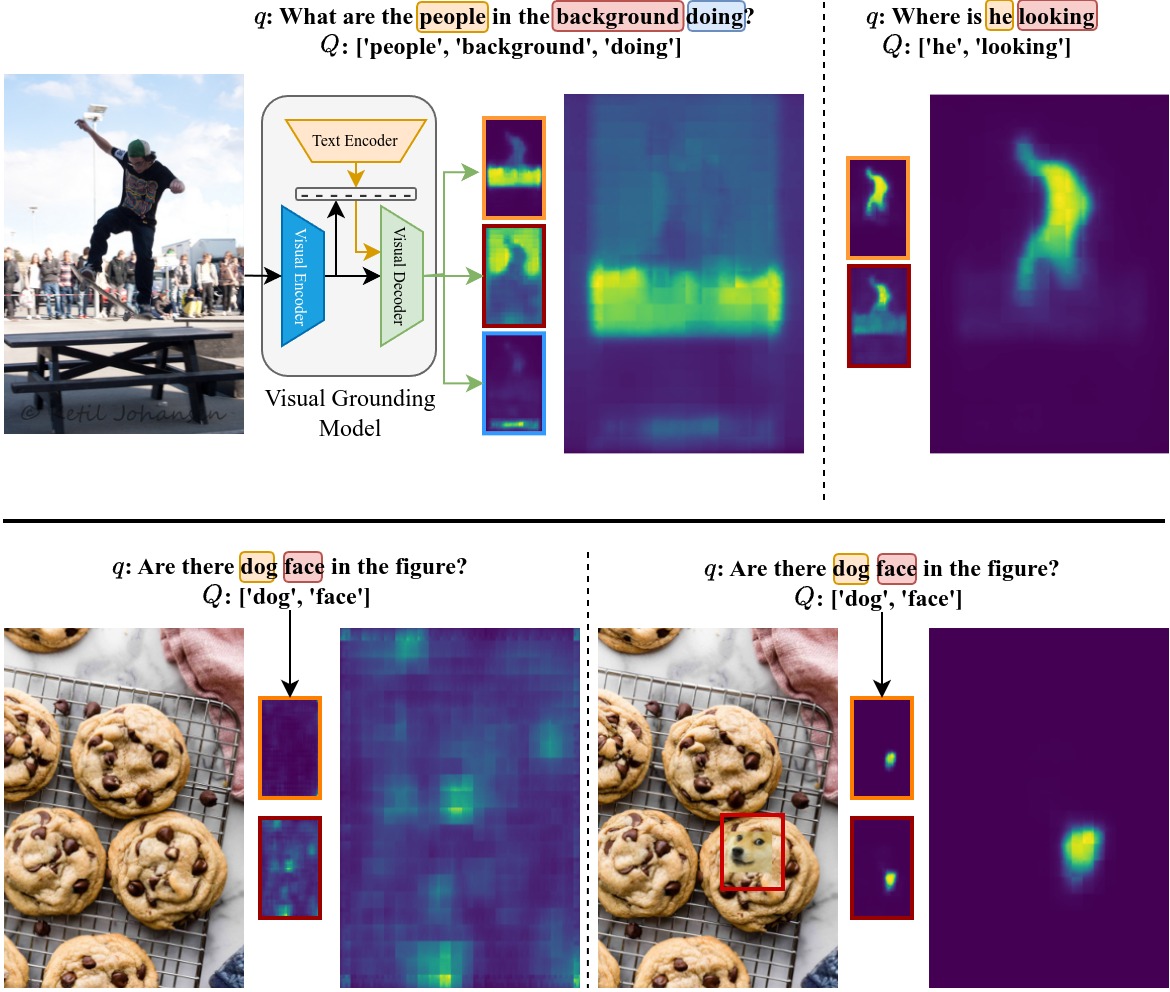}
    \caption{
        \textbf{Illustration of the query-conditioned relevance map.} For the same image (top row), different questions lead to distinct relevance regions aligned with the extracted query terms. Conversely, for the same question (bottom row), the relevance map varies with the image content, localizing evidence that matches the queried concept.
    }
    \label{fig:r_map}
\end{figure}

        \noindent\textbf{Extracting query terms.}
            We first transform the free-form question $q$ into a set of explicit visual queries $Q = \{q_1, q_2, \dots, q_m\}$. Directly using $q$ is often suboptimal, as natural questions may include abstract or relational expressions (e.g., ``Is the person hungry?'') that are difficult for vision encoders to interpret. To bridge this gap, a large language model $f_{\text{LLM}}$ reformulates $q$ into visually grounded queries $Q = f_{\text{LLM}}(q)$, where each $q_i$ denotes a concrete, visually identifiable concept such as an object or attribute. For example, for the question ``What is the man wearing on his feet?'', the $f_{\text{LLM}}$ generates queries like ``shoes'', ``feet'', and ``clothing''. These queries serve as explicit textual anchors that guide vision–language alignment, as illustrated in Fig.~\ref{fig:r_map}.
        
        \noindent\textbf{Computing query-conditioned regions.}
            Given the extracted queries $Q$, each $q_i$ is used to localize its corresponding visual evidence within $x$. We employ a CLIP-based visual grounding model $f_{\text{g}}$, where each text query $q_i$ guides the decoder to produce a spatial relevance map $R(x \mid q_i) = f_{\text{g}}(x, q_i) \in [0,1]^{H \times W}$, representing pixel-wise relevance to $q_i$. This retrieval-like grounding process aligns textual concepts with their spatial counterparts.
                \begin{equation}
                    \label{eq:r_map}
                    r(x \mid q) = \frac{1}{m} \sum_{q_i \in Q} R(x \mid q_i), \quad Q = f_{\text{LLM}}(q).
                \end{equation}
            Averaging across all query terms ensures consistent activation of semantically related regions while suppressing spurious responses as defined in Eq.~\eqref{eq:r_map}. The resulting map $r(x \mid q)$ provides the spatial foundation for evaluating how effectively the LVLM leverages relevant evidence in the subsequent stages of \textit{DnR}.

    \subsection{Question-Conditioned Utilization}
        After obtaining $r(x \mid q)$, the next objective is to assess how effectively the model utilizes the critical visual regions indicated by $r(x \mid q)$. The \textit{question-conditioned utilization} $U_q(x)$ quantifies how the model’s prediction varies when these regions, either highly relevant (\textit{Top-$k$}) or irrelevant (\textit{Bottom-$k$}), are perturbed. A higher $U_q(x)$ indicates that the model responds sensitively to question-critical evidence while remaining stable against distractive regions, indicating more faithful and evidence-grounded reasoning.
        
        \subsubsection{Relevance-based probabilistic masking}
\begin{figure}[t]
    \centering
    \includegraphics[width=1.0\linewidth]{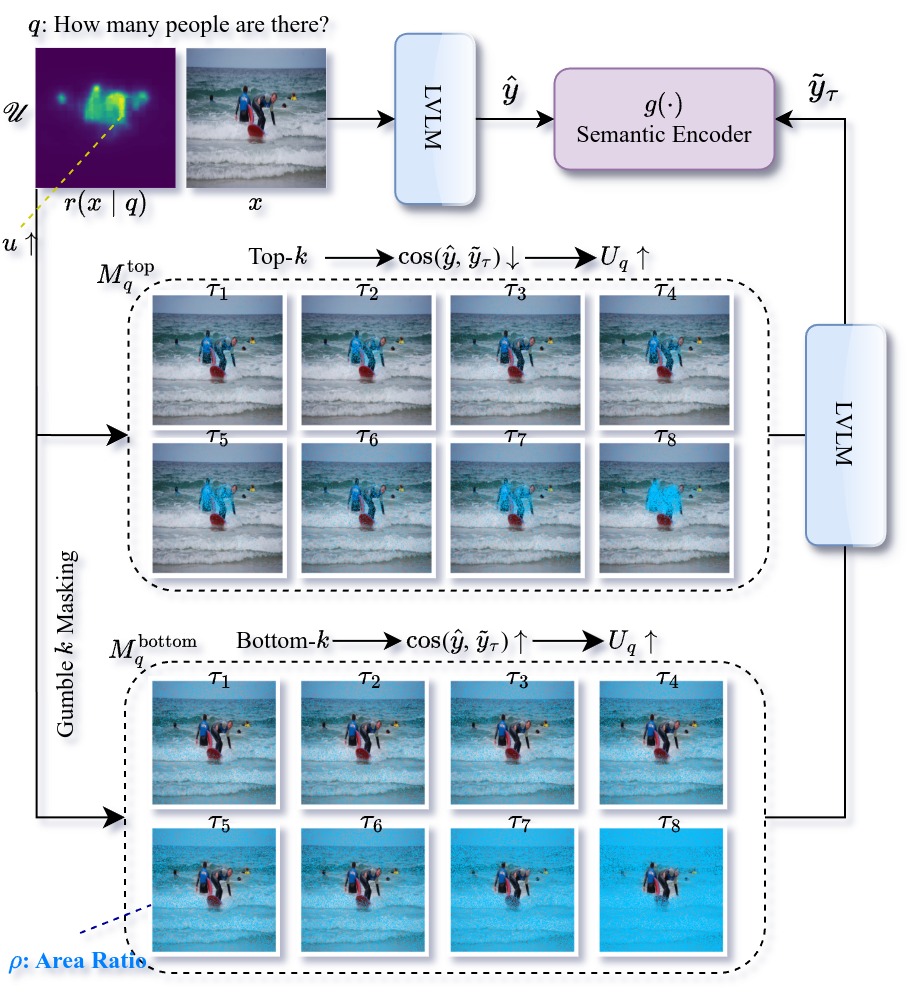}
    \caption{
\textbf{Question-conditioned utilization computation.} Given a question $q$ and image $x$, the relevance map $r(x \mid q)$ guides \textit{Gumbel Top-$k$}/\textit{Bottom-$k$} masking over a ratio $\rho$ of the image. Masked inputs $\tau(x)$ are fed into the LVLM to obtain perturbed predictions $\tilde{y}_\tau$, compared with the original $\hat{y}$ via a semantic encoder $g(\cdot)$, and aggregated with the adaptive factor $\alpha$ to compute the final utilization score $U_q(x)$.
    }
    \label{fig:uq}
\end{figure}

            To quantitatively evaluate how much the model’s prediction depends on the regions identified as relevant, we construct a probability distribution over the spatial regions of the image proportional to their relevance scores. Let $\mathcal{U}(x)$ denote the set of candidate regions within $x$. The probability of each region $u \in \mathcal{U}(x)$ is computed by normalizing its relevance value obtained from the relevance map.
                \begin{equation}
                    \label{eq:r_dist}
                    P(u \mid x, q) =
                    \frac{r(u \mid x, q)}
                         {\sum_{u' \in \mathcal{U}(x)} r(u' \mid x, q)}.
                \end{equation}
            This normalized distribution, Eq.~\eqref{eq:r_dist}, provides a probabilistic weighting over image regions according to their question-conditioned importance. Based on this distribution, two complementary masking strategies are employed, as illustrated in Fig.~\ref{fig:uq}. \textbf{Top-$k$ masking} occludes highly relevant regions (high $P(u \mid x, q)$) to measure degradation in model prediction, while \textbf{Bottom-$k$ masking} occludes less relevant regions (low $P(u \mid x, q)$) to assess prediction stability. Sampling from these two strategies yields a set of $M$ stochastic binary masks, denoted as $\mathcal{M}_q = {\tau_1, \dots, \tau_M}$, each covering approximately a $\rho$ fraction of the image area. To ensure stochastic region selection, we adopt Gumbel-$k$ sampling~\cite{kool2019gumbeltopk, vieillard2020diversesampling}, which enables diverse Top-$k$/Bottom-$k$ mask generation from the relevance distribution.
            \noindent\textbf{Adaptive combination of Top-$k$ and Bottom-$k$ masking.}
                While Top-$k$ and Bottom-$k$ perturbations reveal complementary aspects of model behavior, their relative influence should depend on the sharpness and distribution of the relevance map $r(x \mid q)$. We therefore propose an \textit{adaptive weighting factor} $\alpha \in [0,1]$ per sample by analyzing the information structure of $r(x \mid q)$. Specifically, $\alpha$ is obtained through a function $\alpha = \Phi_{\mathrm{adapt}}\big(r(x \mid q)\big)$ that integrates two cues (\textbf{entropy} and \textbf{contrast}) reflecting the focus and separability of the relevance distribution.
                
                Eq.~\eqref{eq:alpha_adapt} defines this adaptive function as a weighted combination of the normalized entropy $\mathcal{H}_{\mathrm{norm}}$ and contrast $C$:
                \begin{equation}
                    \label{eq:alpha_adapt}
                    \alpha =
                    \frac{
                        \beta_{\mathrm{ent}} \mathcal{H}_{\mathrm{norm}} +
                        \beta_{\mathrm{ctr}} C
                    }{
                        \beta_{\mathrm{ent}} + \beta_{\mathrm{ctr}}
                    }.
                \end{equation}
                A higher $\alpha$ corresponds to a sharper and more distinct relevance map, assigning greater weight to Top-$k$ masking, whereas a lower $\alpha$ emphasizes Bottom-$k$ masking to assess stability under diffuse relevance. This adaptive mechanism allows $U_q(x)$ to dynamically decide which perturbation to emphasize based on the input, focusing on decisive evidence when the relevance map is confident and prioritizing stability when it is uncertain.

        \subsubsection{Semantic deviation measurement.}
            For each mask $\tau \in \mathcal{M}_q$, a masked image $\tau(x)$ is generated by removing the selected regions. Let $\hat{y} = f_{\mathrm{VLM}}(x,q)$ denote the original prediction and $\tilde{y}_\tau = f_{\mathrm{VLM}}(\tau(x), q)$ the masked prediction. Both are projected into an embedding space by a semantic encoder $g(\cdot)$, yielding $z_{\hat{y}} = g(\hat{y})$ and $z_{\tilde{y}_\tau} = g(\tilde{y}_\tau)$.

            The semantic deviation for each mask is defined differently depending on its type (top-$k$ or bottom-$k$).
                \begin{equation}
                    \label{eq:semantic_shift}
                    d_{\tau}(\hat{y}, \tilde{y}_\tau) = 
                        \begin{cases} 1 - \cos\!\big(z_{\hat{y}}, z_{\tilde{y}_\tau}\big), & \text{if } \tau \in \mathcal{M}_q^{\text{top}}, \\[4pt] \cos\!\big(z_{\hat{y}}, z_{\tilde{y}_\tau}\big), & \text{if } \tau \in \mathcal{M}_q^{\text{bottom}}. \end{cases}
                \end{equation}
            As shown in Eq.~\eqref{eq:semantic_shift}, Top-$k$ masking measures the drop in semantic similarity after masking highly relevant regions, whereas Bottom-$k$ masking quantifies the stability of predictions when irrelevant regions are masked.

        \subsubsection{Utilization score.}
            Building upon the semantic deviations defined in Eq.~\eqref{eq:semantic_shift}, the final utilization score $U_q(x)$ aggregates the semantic deviations from both Top-$k$ and Bottom-$k$ perturbations through the adaptive factor $\alpha$. It quantifies how the model’s prediction changes when query-relevant or irrelevant regions are masked, providing a balanced measure of evidence dependence and robustness.
                \begin{align}
                    \label{eq:utilization_score}
                    U_q(x) &=
                    \alpha \cdot
                    \mathbb{E}_{\tau \in \mathcal{M}_q^{\text{top}}}
                    [d_{\tau}(\hat{y}, \tilde{y}_{\tau})] \nonumber \\
                    &\quad+
                    (1 - \alpha) \cdot
                    \mathbb{E}_{\tau \in \mathcal{M}_q^{\text{bottom}}}
                    [d_{\tau}(\hat{y}, \tilde{y}_{\tau})].
                \end{align}      
            In Eq.~\eqref{eq:utilization_score}, the expectation operator $\mathbb{E}_{\tau \in \mathcal{M}_q^{\text{top}}}[\cdot]$ denotes the mean semantic deviation over all masks in each subset. Owing to the complementary definitions in Eq.~\eqref{eq:semantic_shift}, a higher $U_q(x)$ consistently indicates a more faithful and visually grounded reasoning process. Specifically, the Top-$k$ term captures how \textbf{strongly the model relies} on question-critical regions, where larger deviations imply stronger evidence dependence, whereas the Bottom-$k$ term measures how \textbf{stable the prediction remains} when irrelevant regions are perturbed. Their contributions are adaptively balanced by the factor $\alpha$, yielding a unified measure of both dependence and robustness.

    \subsection{Expert Selection and Incorporation}
        \begin{figure}[t]
            \centering
            \includegraphics[width=1.0\linewidth]{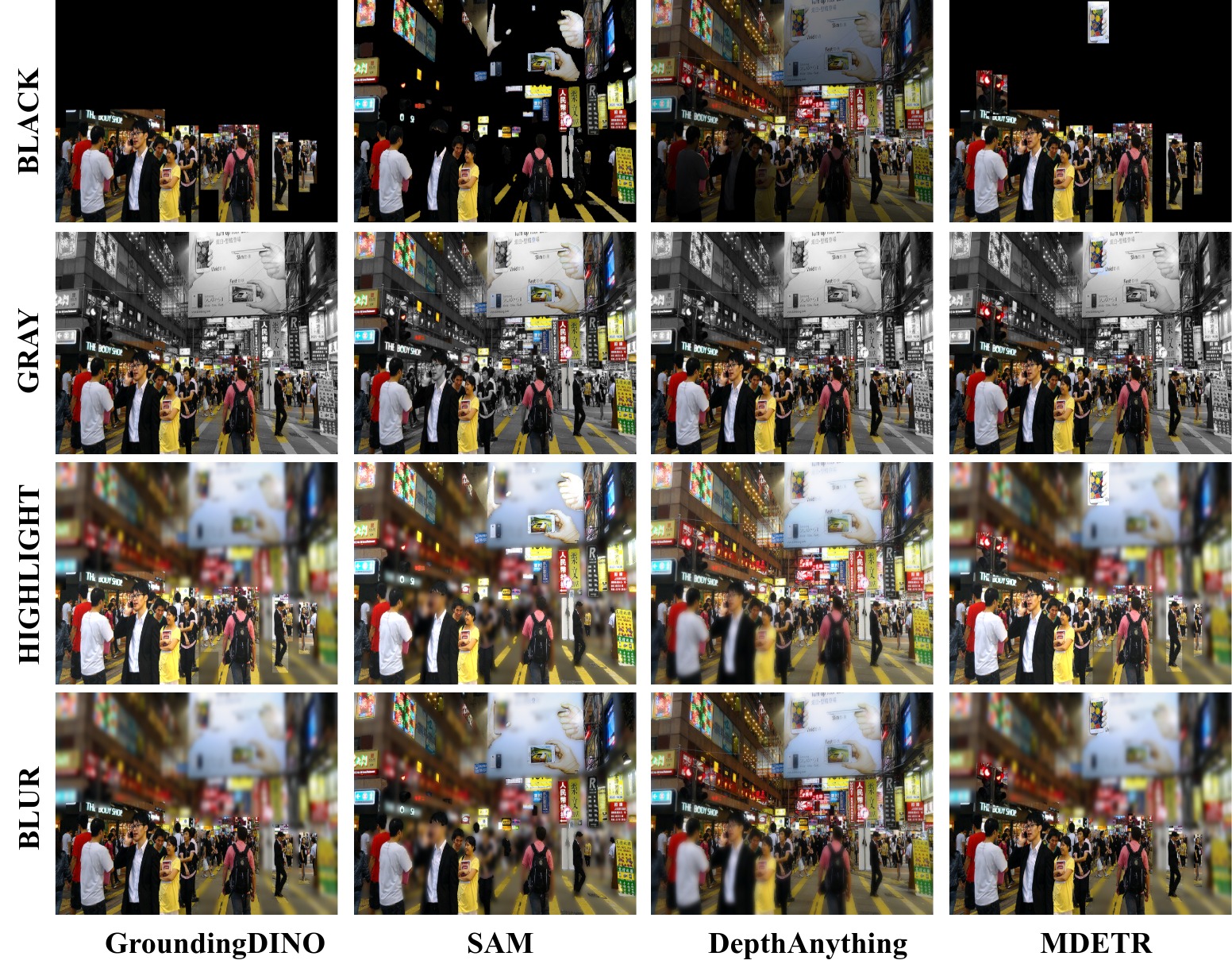}
            \caption{
                Comparison of rendering strategies across different experts. Each column corresponds to an experts, and each row represents a rendering style.
            }
            \label{fig:rendering}
        \end{figure}

         \textit{DnR} leverages $U_q$ as a quantitative measure to guide the selection of visual experts. Given multiple candidates $\{h_1, h_2, \dots, h_K\}$, each expert generates structured outputs such as bounding boxes, masks, or depth maps, which the LVLM cannot directly process. To make these outputs compatible, they are rendered onto the original image to produce $\hat{x}^{(j)} = R(x, h_j)$, forming rendered visual evidence. This rendering converts the expert’s structural predictions into visual cues (e.g., gray, blur, or highlight regions) that the model can process through its existing vision encoder without any architectural modification (Fig.~\ref{fig:rendering}).
            \begin{equation}
                \label{eq:refinement}
                \tilde{y}^{(j)} = f_{\mathrm{VLM}}(\hat{x}^{(j)}, q) = f_{\mathrm{VLM}}(R(x, h_j), q)
            \end{equation}
        Each $\tilde{y}^{(j)}$ represents the model’s refined response obtained by re-querying $f_{\mathrm{VLM}}$  with the rendered input $\hat{x}^{(j)} = R(x, h_j)$, where the rendering $R(\cdot)$ visually encodes the expert’s structural output onto the original image to serve as explicit visual evidence.

        Subsequently, the framework computes the query-conditioned utilization $U_q^{(j)}$ for each refined response $\tilde{y}^{(j)}$ and compares it to the baseline utilization $U_q^{\mathrm{base}}$ obtained from the initial draft $\hat{y}$ (Fig.~\ref{fig:dnr}). For each expert-rendered input $\hat{x}^{(j)}$ and its prediction $\tilde{y}^{(j)}$, the utilization score $U_q^{(j)}$ is re-computed by applying the original masks $\tau \in \mathcal{M}_q$ to $\hat{x}^{(j)}$ and using $\tilde{y}^{(j)}$ as the new baseline for Eq.~\eqref{eq:semantic_shift}.
        The expert that yields the largest improvement is then selected:
            \begin{equation}
                \label{eq:expert_selection}
                j^* = \arg\max_j \big(U_q^{(j)} - U_q^{\mathrm{base}}\big)_+,
            \end{equation}
        If no expert increases $U_q(x)$ beyond $U_q^{\mathrm{base}}$, the system concludes that the draft already captures sufficient visual grounding and skips further refinement, where $(\cdot)_+$ ensures that only positive improvements over the baseline are considered.

    \subsection{Learned Expert Selection}
        Running all $K$ experts for Eq.~\eqref{eq:expert_selection} is computationally expensive. To make \textit{DnR} practical, a lightweight \textbf{selector} network $S_\theta (j \mid s)$ is trained to predict the optimal expert $j^*$ as defined in Eq.~\eqref{eq:expert_selection} directly from the state $s$.
        
        The state $s$ represents the query-conditioned context observed before expert invocation:
            \begin{equation}
                \label{eq:state_def}
                s = (x,\, Q,\, \hat{y},\, r(x \mid q)),
            \end{equation}
            where $x$ is the image, $Q$ the query set, $\hat{y}$ the draft prediction, and $r(x \mid q)$ the relevance map. The selector $S_\theta$ is trained with the loss
                \begin{equation}
                    \label{eq:selector_loss}
                    \mathcal{L} =
                    - \mathbb{E}_{s \sim \mathcal{D}}
                    \left[ \log S_\theta(j^* \mid s) \right].
                \end{equation}
                
        This setup enables scalable expert coordination as the number of experts increases, where direct evaluation becomes linearly expensive since each candidate requires separate rendering and utilization computation. Leveraging the utilization difference $(U_q^{(j)} - U_q^{\mathrm{base}})$ defined in \textit{DnR}, the selector learns an approximate rule that converts the refinement process from exhaustive search to direct decision.
        
            \begin{table*}[t]
            \centering
            \resizebox{\textwidth}{!}{
                \begin{tabular}{l|ccccc|ccc|cc|ccc|ccc}
                    \toprule
                    \multirow{2}{*}{\textbf{VLM Backbone}} 
                        & \multicolumn{5}{c|}{\textbf{VQA}} 
                        & \multicolumn{3}{c|}{\textbf{Image Captioning}} 
                        & \multicolumn{2}{c|}{\textbf{Visual Reasoning}} 
                        & \multicolumn{3}{c|}{\textbf{Knowledge VQA}} 
                        & \multicolumn{3}{c}{\textbf{Comprehensive Benchmarks}} \\
                    \cmidrule(lr){2-6} \cmidrule(lr){7-9} \cmidrule(lr){10-11} \cmidrule(lr){12-14} \cmidrule(lr){15-17}
                        & VQAv2~\cite{goyal2017vqav2} & GQA~\cite{hudson2019gqa} & VizWiz~\cite{gurari2018vizwiz} & TextVQA~\cite{singh2019textvqa} & OCR-VQA~\cite{mishra2019ocrvqa} 
                        & COCO~\cite{lin2014coco} & NoCaps~\cite{agrawal2019nocaps} & Flickr~\cite{plummer2015flickr30k}
                        & VCR~\cite{zellers2019vcr} & VSR~\cite{xie2023vsr} 
                        & OK-VQA~\cite{marino2019okvqa} & A-OKVQA~\cite{schwenk2022aokvqa} & ScienceQA~\cite{lu2022scienceqa} 
                        & MME~\cite{fu2024mme} & MMBench~\cite{liu2024mmbench} & SEED-Bench~\cite{li2024seedbench} \\
                    
                    \midrule
                    
                    IDEFICS~\cite{laurencon2024idefics2} (Draft / DnR)
                        & \cellcolor{Green!15}37.8 / \textbf{47.85} & \cellcolor{Green!15}24.1 / \textbf{25.5} & \textbf{43.33} / 36.67 & \cellcolor{Green!15}30.15 / \textbf{30.32} & \cellcolor{Green!15}47.24 / \textbf{47.74} 
                        & \cellcolor{Green!15}135.66 / \textbf{137.7} & \cellcolor{Green!15}106.53 / \textbf{108.5} & \cellcolor{Green!15}74.2 / \textbf{80.4} 
                        & \cellcolor{Green!15}15.58 / \textbf{21.11} & \cellcolor{Green!15}52.76 / \textbf{54.27} 
                        & \cellcolor{Green!15}43.5 / \textbf{44.4} & \cellcolor{Green!15}69 / \textbf{69.5} & \cellcolor{Green!15}43.95 / \textbf{44.02} 
                        & \cellcolor{Green!15}1392.11 / \textbf{1431.58} & \cellcolor{Green!15}50.01 / \textbf{50.26} & \cellcolor{Green!15}32.16 / \textbf{32.75} \\
                    Revision Rate & 29.8 & 1.5 & 19.8 & 2 & 3.2 & 17.8 & 17 & 18.4 & 18.3 & 2.5 & 2.9 & 0.2 & 0.2 & 0.6 & 0.4 & 0.3 \\
                    Correction / Degradation 
                        & 46.2 / 14.3 & 51.3 / 1.1 & 6.3 / 24.1 & 2.3 / 0.3 & 12.5 / 2.1 
                        & - & - & - & 56.3 / 7.1 & 52.9 / 32.3 
                        & 33.3 / 1.4 & 8.6 / 1.9 & 94.3 / 0.4 
                        & 80.3 / 16.6 & 43.8 / 12.2 & 33.3 / 19.6 \\
                    Pearson/Spearman 
                        & 0.143 / 0.064 & \cellcolor{RoyalBlue!15}0.449 / \cellcolor{RoyalBlue!15}0.364 & \cellcolor{RoyalBlue!15}0.248 / \cellcolor{RoyalBlue!15}0.259 & 0.026 / 0.073 & 0.129 / \cellcolor{RoyalBlue!15}0.222 
                        & - & - & - & \cellcolor{RoyalBlue!15}0.38 / \cellcolor{RoyalBlue!15}0.421 & 0.158 / 0.173 
                        & \cellcolor{RoyalBlue!15}0.351 / 0.210 & 0.1 / 0.166 & \cellcolor{RoyalBlue!15}0.351 / \cellcolor{RoyalBlue!15}0.277 
                        & 0.12 / 0.148 & 0.12 / 0.06 & \cellcolor{RoyalBlue!15}0.354 / \cellcolor{RoyalBlue!15}0.375 \\

                    \midrule
    
                    InstructBLIP~\cite{dai2023instructblip}
                        & \cellcolor{Green!15}76.4 / \textbf{77.75} & \cellcolor{Green!15}38.24 / \textbf{39.77} & \cellcolor{Green!15}37.83 / \textbf{38.67} & \cellcolor{Green!15}52.43 / \textbf{54.1} & \cellcolor{Green!15}80.4 / \textbf{81.91} 
                        & \cellcolor{Green!15}114.31 / \textbf{118.7} & \cellcolor{Green!15}107.8 / \textbf{109.1} & \cellcolor{Green!15}72.7 / \textbf{79.8} 
                        & \textbf{13.07} / 12.56 & \cellcolor{Green!15}61.31 / \textbf{61.81} 
                        & \cellcolor{Green!15}49.2 / \textbf{50.2} & \cellcolor{Green!15}79.11 / \textbf{81.09} & \cellcolor{Green!15}50.5 / \textbf{52.5} 
                        & \cellcolor{Green!15}1294.74 / \textbf{1295.31} & \cellcolor{Green!15}51.84 / \textbf{52.89} & \cellcolor{Green!15}51.46 / \textbf{53.8} \\
                    Revision Rate 
                        & 7.2 & 13 & 24.2 & 2.1 & 1.2 & 26.2 & 31.1 & 36 & 4.1 & 1.2 & 4.5 & 1.8 & 1.2 & 0.1 & 1.6 & 3.2 \\
                    Correction / Degradation 
                        & 9.1 / 5.4 & 35.3 / 1.2 & 16.1 / 2 & 52.3 / 25.25 & 24.5 / 3.4 
                        & - & - & - & 13.2 / 21.3 & 59.9 / 33.1 
                        & 39.5 / 18.4 & 42.9 / 28.6 & 77.6 / 4.3 
                        & 92.1 / 4.3 & 55.3 / 4.4 & 36.4 / 9.1 \\
                    Pearson/Spearman 
                        & \cellcolor{RoyalBlue!15}0.243 / -0.024 & \cellcolor{RoyalBlue!15}0.290 / \cellcolor{RoyalBlue!15}0.286 & \cellcolor{RoyalBlue!15}0.290 / \cellcolor{RoyalBlue!15}0.286 & 0.066 / 0.076 & \cellcolor{RoyalBlue!15}0.426 / \cellcolor{RoyalBlue!15}0.415 
                        & - & - & - & \cellcolor{RoyalBlue!15}0.2 / \cellcolor{RoyalBlue!15}0.204 & \cellcolor{RoyalBlue!15}0.444 / \cellcolor{RoyalBlue!15}0.361 
                        & 0.168 / 0.09 & 0.136 / 0.104 & \cellcolor{RoyalBlue!15}0.608 / \cellcolor{RoyalBlue!15}0.421 
                        & 0.152 / 0.128 & 0.01 / 0.02 & 0.062 / 0.145 \\

                    \midrule
                    
                    MiniGPTv2~\cite{zhu2023minigptv2}
                        & \cellcolor{Green!15}32.6 / \textbf{34.1} & \cellcolor{Green!15}25.3 / \textbf{27.6} & \cellcolor{Green!15}59.17 / \textbf{60.67} & \cellcolor{Green!15}36.67 / \textbf{36.68} & \cellcolor{Green!15}56.78 / \textbf{58.79} 
                        & - & - & - 
                        & \cellcolor{Green!15}13.07 / \textbf{15.58} & \cellcolor{Green!15}40.2 / \textbf{43.72} 
                        & \cellcolor{Green!15}18.1 / \textbf{19.8} & \cellcolor{Green!15}38.58 / \textbf{41.01} & \cellcolor{Green!15}28.77 / \textbf{29.41} 
                        & \cellcolor{Green!15}878.95 / \textbf{910.53} & \cellcolor{Green!15}37.63 / \textbf{39.21} & \cellcolor{Green!15}29.82 / \textbf{31.58} \\
                    Revision Rate 
                        & 17.5 & 3.1 & 3.8 & 15.5 & 1.8 & - & - & - & 7.2 & 0.8 & 4.5 & 6.8 & 0.2 & 0.2 & 0.4 & 7.6 \\
                    Correction / Degradation 
                        & 12.9 / 2.9 & 8.3 / 0.5 & 41.2 / 6.7 & 1.6 / 0.1 & 11.1 / 0.1 
                        & - & - & - & 49.9 / 3.4 & 65.1 / 20.8 
                        & 11.1 / 0.1 & 14.8 / 7.2 & 94.3 / 0.3 
                        & 50.32 / 12.1 & 66.7 / 1.2 & 30.8 / 15.7 \\
                    Pearson/Spearman 
                        & 0.026 / 0.051 & 0.194 / \cellcolor{RoyalBlue!15}0.378 & \cellcolor{RoyalBlue!15}0.304 / \cellcolor{RoyalBlue!15}0.234 & \cellcolor{RoyalBlue!15}0.492 / 0.129 & 0.065 / 0.124 
                        & - & - & - & \cellcolor{RoyalBlue!15}0.292 / \cellcolor{RoyalBlue!15}0.224 & 0.155 / 0.165 
                        & \cellcolor{RoyalBlue!15}0.206 / 0.168 & 0.108 / 0.122 & \cellcolor{RoyalBlue!15}0.719 / \cellcolor{RoyalBlue!15}0.807 
                        & \cellcolor{RoyalBlue!15}0.347 / \cellcolor{RoyalBlue!15}0.35 & \cellcolor{RoyalBlue!15}0.422 / \cellcolor{RoyalBlue!15}0.360 & 0.155 / 0.131 \\
                        
                    \midrule
    
                    LLaVA~1.6~\cite{liu2024llava16}
                        & \cellcolor{Green!15}80.9 / \textbf{82.81} & \cellcolor{Green!15}61.5 / \textbf{64.2} & \cellcolor{Green!15}76.83 / \textbf{76.99} & \cellcolor{Green!15}64.49 / \textbf{64.59} & \cellcolor{Green!15}73.37 / \textbf{74.87} 
                        & \cellcolor{Green!15}126.54 / \textbf{138.6} & \cellcolor{Green!15}70.18 / \textbf{75.5} & \cellcolor{Green!15}77.9 / \textbf{79.6} 
                        & \cellcolor{Green!15}18.59 / \textbf{18.69} & \cellcolor{Green!15}64.82 / \textbf{65.83} 
                        & \cellcolor{Green!15}55.1 / \textbf{56.1} & 73.5 / 73.5 & \cellcolor{Green!15}72.9 / \textbf{73.8} 
                        & \cellcolor{Green!15}1694.74 / \textbf{1721.05} & \cellcolor{Green!15}76.32 / \textbf{77.89} & \cellcolor{Green!15}66.08 / \textbf{66.67} \\
                    Revision Rate 
                        & 5.1 & 4.5 & 1.8 & 1 & 3.9 & 22.8 & 37.1 & 23 & 0.5 & 0.2 & 2 & 0.8 & 1.4 & 0.4 & 0.5 & 0.3 \\
                    Correction / Degradation 
                        & 66.7 / 9.2 & 33.3 / 3.1 & 14.3 / 2.1 & 25 / 0.7 & 66.7 / 16.7 
                        & - & - & - & 25.3 / 11.4 & 95.3 / 0.3 
                        & 37.5 / 24.9 & 33.3 / 7.8 & 49.8 / 0.4 
                        & 98.3 / 0.6 & 23.3 / 7.3 & 95.5 / 0.4 \\
                    Pearson/Spearman 
                        & \cellcolor{RoyalBlue!15}0.509 / \cellcolor{RoyalBlue!15}0.639 & \cellcolor{RoyalBlue!15}0.288 / \cellcolor{RoyalBlue!15}0.328 & 0.156 / 0.154 & 0.155 / \cellcolor{RoyalBlue!15}0.231 & \cellcolor{RoyalBlue!15}0.626 / \cellcolor{RoyalBlue!15}0.849 
                        & - & - & - & \cellcolor{RoyalBlue!15}0.772 / \cellcolor{RoyalBlue!15}0.778 & \cellcolor{RoyalBlue!15}0.272 / \cellcolor{RoyalBlue!15}0.272 
                        & \cellcolor{RoyalBlue!15}0.297 / 0.161 & 0.110 / 0.128 & \cellcolor{RoyalBlue!15}0.454 / \cellcolor{RoyalBlue!15}0.459 
                        & \cellcolor{RoyalBlue!15}0.270 / \cellcolor{RoyalBlue!15}0.307 & 0.109 / 0.116 & 0.112 / 0.01 \\
                        
                    \midrule
    
                    PaliGemma~\cite{desai2024paligemma}
                        & \cellcolor{Green!15}73.2 / \textbf{75.2} & \cellcolor{Green!15}58.3 / \textbf{59.9} & \cellcolor{Green!15}79.17 / \textbf{80.17} & \cellcolor{Green!15}65.83 / \textbf{66.5} & \cellcolor{Green!15}68.34 / \textbf{70.35} 
                        & \cellcolor{Green!15}137.99 / \textbf{144.7} & \cellcolor{Green!15}83.81 / \textbf{85.99} & \cellcolor{Green!15}92.9 / \textbf{101.9} 
                        & \cellcolor{Green!15}13.57 / \textbf{15.53} & \cellcolor{Green!15}66.83 / \textbf{68.84} 
                        & \cellcolor{Green!15}57.2 / \textbf{58.5} & \cellcolor{Green!15}85.2 / \textbf{85.5} & \cellcolor{Green!15}89.1 / \textbf{89.7} 
                        & \cellcolor{Green!15}1434.21 / \textbf{1444.74} & \cellcolor{Green!15}70.09 / \textbf{71.05} & \cellcolor{Green!15}57.89 / \textbf{59.65} \\
                    Revision Rate 
                        & 9.2 & 1.5 & 2.8 & 1.2 & 3.5 & 22.6 & 38.2 & 53.2 & 2.2 & 0.9 & 3.2 & 1 & 0.5 & 0.2 & 0.5 & 4.1 \\
                    Correction / Degradation 
                        & 36.4 / 27.3 & 16.7 / 0.8 & 82.1 / 2.3 & 42.3 / 21.3 & 28.6 / 0.5 
                        & - & - & - & 43.4 / 5.4 & 75.5 / 11.1 
                        & 60.15 / 23.1 & 12.3 / 0.3 & 92.1 / 1.3 
                        & 51.1 / 12.5 & 23.5 / 2.1 & 64.3 / 31.7 \\
                    Pearson/Spearman 
                        & \cellcolor{RoyalBlue!15}0.292 / \cellcolor{RoyalBlue!15}0.291 & \cellcolor{RoyalBlue!15}0.276 / \cellcolor{RoyalBlue!15}0.399 & \cellcolor{RoyalBlue!15}0.307 / \cellcolor{RoyalBlue!15}0.251 & 0.136 / 0.03 & \cellcolor{RoyalBlue!15}0.321 / \cellcolor{RoyalBlue!15}0.329 
                        & - & - & - & \cellcolor{RoyalBlue!15}0.253 / \cellcolor{RoyalBlue!15}0.265 & \cellcolor{RoyalBlue!15}0.341 / \cellcolor{RoyalBlue!15}0.275 
                        & 0.165 / 0.128 & \cellcolor{RoyalBlue!15}0.441 / \cellcolor{RoyalBlue!15}0.454 & \cellcolor{RoyalBlue!15}0.803 / \cellcolor{RoyalBlue!15}0.853 
                        & 0.08 / 0.07 & 0.09 / 0.1 & 0.129 / 0.191 \\
                        
                    \midrule
    
                    CogVLM~\cite{hong2024cogvlm}
                        & \cellcolor{Green!15}82.05 / \textbf{82.85} & \cellcolor{Green!15}56.13 / \textbf{57.74} & \cellcolor{Green!15}48.5 / \textbf{50.33} & \cellcolor{Green!15}68.51 / \textbf{69.68} & 82.91 / 82.91 
                        & \cellcolor{Green!15}86.57 / \textbf{94.8} & \cellcolor{Green!15}80.3 / \textbf{81.7} & \cellcolor{Green!15}67.7 / \textbf{70.5} 
                        & \cellcolor{Green!15}12.56 / \textbf{13.29} & \cellcolor{Green!15}62.81 / \textbf{65.33} 
                        & 58.6 / 58.6 & 84.5 / 84.5 & \cellcolor{Green!15}61.5 / \textbf{62.22} 
                        & \cellcolor{Green!15}1384.21 / \textbf{1423.68} & \cellcolor{Green!15}76.84 / \textbf{77.89} & \cellcolor{Green!15}58.48 / \textbf{59.06} \\
                    Revision Rate 
                        & 5.1 & 2.5 & 4.8 & 2.30 & 0.20 
                        & 13.4 & 11.1 & 30.6 
                        & 2.5 & 0.5 
                        & 0.8 & 0.7 & 1 
                        & 0.6 & 0.3 & 0.3 \\
                    Correction / Degradation 
                        & 66.7 / 2.3 & 32.2 / 1.3 & 5.7 / 0.4 & 42.9 / 28.6 & 1.3 / 0.1 
                        & - & - & - 
                        & 56.3 / 19.4 & 99.7 / 0.1 
                        & 33.1 / 4.6 & 3.4 / 2.1 & 75.43 / 21.33 
                        & 66.6 / 31.1 & 70.1 / 2.9 & 98.1 / 0.3 \\
                    Pearson / Spearman 
                        & \cellcolor{RoyalBlue!15}0.684 / \cellcolor{RoyalBlue!15}0.735 & \cellcolor{RoyalBlue!15}0.223 / \cellcolor{RoyalBlue!15}0.357 & \cellcolor{RoyalBlue!15}0.220 / 0.136 & \cellcolor{RoyalBlue!15}0.318 / 0.11 & \cellcolor{RoyalBlue!15}0.263 / \cellcolor{RoyalBlue!15}0.28 
                        & - & - & - 
                        & 0.110 / 0.117 & \cellcolor{RoyalBlue!15}0.458 / \cellcolor{RoyalBlue!15}0.456 
                        & \cellcolor{RoyalBlue!15}0.500 / \cellcolor{RoyalBlue!15}0.623 & 0.163 / 0.14 & \cellcolor{RoyalBlue!15}0.499 / \cellcolor{RoyalBlue!15}0.412 
                        & \cellcolor{RoyalBlue!15}0.420 / \cellcolor{RoyalBlue!15}0.470 & 0.01 / 0.03 & 0.153 / \cellcolor{RoyalBlue!15}0.227 \\
                        
                    \midrule
                    
                    Qwen2.5-VL~\cite{qwen2025}
                        & \cellcolor{Green!15}83.95 / \textbf{85.45} & \cellcolor{Green!15}57.02 / \textbf{58.31} & \cellcolor{Green!15}73.01 / \textbf{73.83} & \cellcolor{Green!15}83.92 / \textbf{84.25} & \cellcolor{Green!15}72.36 / \textbf{74.86} 
                        & \cellcolor{Green!15}61.96 / \textbf{64.7} & \cellcolor{Green!15}71.8 / \textbf{74.2} & \textbf{57.3} / 55.9 
                        & \cellcolor{Green!15}30.65 / \textbf{31.66} & \cellcolor{Green!15}79.9 / \textbf{80.4} 
                        & \cellcolor{Green!15}58.2 / \textbf{58.3} & \cellcolor{Green!15}74.4 / \textbf{76.4} & \cellcolor{Green!15}86.5 / \textbf{87.1} 
                        & \cellcolor{Green!15}2268.42 / \textbf{2276.32} & \cellcolor{Green!15}86.05 / \textbf{86.84} & \cellcolor{Green!15}80.12 / \textbf{81.29} \\
                    Revision Rate 
                        & 10.6 & 8.8 & 1.2 & 0.20 & 2.5 & 27.6 & 37.5 & 42.1 & 5.5 & 0.2 & 3.8 & 1.5 & 0.4 & 0.3 & 1.2 & 0.5 \\
                    Correction / Degradation 
                        & 40 / 4.3 & 16.7 / 8.3 & 41.3 / 1.0 & 33.3 / 5.7 & 89.3 / 3.4 
                        & - & - & - & 45.5 / 18.2 & 93.9 / 0.3 
                        & 13.3 / 6.7 & 33.2 / 16.8 & 82.4 / 5.3 
                        & 65.6 / 32.1 & 55.5 / 42.3 & 29.4 / 17.6 \\
                    Pearson/Spearman 
                        & \cellcolor{RoyalBlue!15}0.389 / \cellcolor{RoyalBlue!15}0.509 & 0.136 / \cellcolor{RoyalBlue!15}0.222 & 0.19 / \cellcolor{RoyalBlue!15}0.224 & \cellcolor{RoyalBlue!15}0.230 / 0.112 & \cellcolor{RoyalBlue!15}0.499 / \cellcolor{RoyalBlue!15}0.425 
                        & - & - & - & \cellcolor{RoyalBlue!15}0.637 / \cellcolor{RoyalBlue!15}0.631 & \cellcolor{RoyalBlue!15}0.25 / \cellcolor{RoyalBlue!15}0.25 
                        & 0.02 / 0.023 & \cellcolor{RoyalBlue!15}0.43 / \cellcolor{RoyalBlue!15}0.467 & \cellcolor{RoyalBlue!15}0.644 / \cellcolor{RoyalBlue!15}0.703 
                        & \cellcolor{RoyalBlue!15}0.5 / \cellcolor{RoyalBlue!15}0.5 & 0.188 / 0.083 & 0.170 / 0.07 \\

                    \bottomrule
                \end{tabular}
            }
            \caption{                
                Comprehensive evaluation of diverse LVLMs on 16 multimodal benchmarks. Each cell reports \textit{Draft / DnR} performance. Additional rows show revision rates (percentage of refined), correction vs.\ degradation (\textit{False→True / True→False}) transitions, and correlation coefficients (Pearson / Spearman) measuring alignment between confidence or utilization scores and accuracy changes. Cells in \textcolor{green!50!black}{green} denote improvements after refinement, and those in \textcolor{blue!60!black}{blue} indicate strong correlations (\(r > 0.2\)).
            }
            \label{tab:vlm_task_dataset}
        \end{table*}
        
\section{Experiments}
    We evaluate \textit{DnR} to validate the proposed \(U_q\) and its role in multimodal reasoning. The framework performs a single-step selection by choosing the candidate that maximizes \(U_q\), and experiments examine how this choice influences accuracy, reduces hallucination, and promotes evidence-grounded reasoning.
    
    We use LLaMA-3-70B~\cite{touvron2024llama3} as the language backbone \(f_{\text{LLM}}\) for query decomposition, CLIP-L/14~\cite{radford2021clip} and SentenceTransformer all-MiniLM-L6-v2~\cite{reimers2019sentencebert} as the semantic encoder \(g(\cdot)\), and CLIPSeg~\cite{luddecke2022clipseg} as the visual grounding model \(f_{\mathrm{g}}\). Four visual experts (GroundingDINO~\cite{liu2023groundingdino}, SAM~\cite{kirillov2023segment}, DepthAnything~\cite{yang2024depthanything}, and mDETR~\cite{kamath2021mdetr}) are chosen from distinct backbone families to provide complementary cues. For the utilization computation, we sample \(M = 16\) stochastic masks per image using Gumbel-$k$ sampling, with the masking ratio \(\rho\) set to 0.25 for Top-\(k\), 0.75 for Bottom-\(k\), and adaptively adjusted in Hybrid mode.

    \subsection{Comprehensive Evaluation}
        We evaluate \textit{DnR} across five categories: VQA, image captioning, visual reasoning, knowledge VQA, and comprehensive benchmarks, each examining a distinct aspect of multimodal reasoning with subsets. All experiments use the \textit{GRAY} rendering strategy and \textit{Hybrid} masking mode for consistency. The utilization metric \(U_q\) acts as a unified indicator connecting visual evidence to model behavior, and its validity is tested through Pearson and Spearman correlations between utilization scores and accuracy gains.
        
        \subsubsection{VQA}
            \textit{DnR} improves visual reasoning across diverse LVLMs, but its effect varies with each model’s baseline grounding. Models with weaker visual reliance, such as IDEFICS~\cite{laurencon2024idefics2} and MiniGPT-v2~\cite{zhu2023minigptv2}, show higher revision rates ($\approx$8.3–11.3\%) and larger gains (+1.4–3.0\%), indicating that \textit{DnR} supplies missing visual feedback. Stronger systems like LLaVA 1.6~\cite{liu2024llava16}, CogVLM~\cite{hong2024cogvlm}, and Qwen2.5-VL~\cite{qwen2025} revise less often ($\approx$3.0–4.6\%) yet still improve (+1.1–1.5\%), suggesting a stabilizing effect that refines already coherent reasoning. This contrast shows that \textit{DnR} enhances consistency while moderating excessive or insufficient visual dependence during inference.

            \textit{DnR} scales with each task’s visual demand. Perception-focused benchmarks such as VQAv2~\cite{goyal2017vqav2} and VizWiz~\cite{gurari2018vizwiz} show larger gains (1.8–2.7\%), while text-centric tasks like OCR-VQA~\cite{mishra2019ocrvqa} and TextVQA~\cite{singh2019textvqa} show smaller but consistent improvements (1.4–1.6\%). An exception appears in IDEFICS–VizWiz, the drop occurs because DnR pushes the model to answer cases it previously avoided by saying “unanswerable,” and its limited capability on these queries naturally leads to more mistakes. A high correction-to-degradation ratio (31.8–6.5\%) and positive Pearson and Spearman correlations (0.275/0.273) indicate that \textit{DnR} systematically shifts predictions toward more evidence-grounded reasoning rather than random variation.
                                    
        \subsubsection{Image Captioning}
            In image captioning, where no explicit $q$ is provided, \textit{DnR} constructs object-centric $Q$ to explore the scene and guide visual reasoning. The strength of improvement depends on each model’s initial captioning bias. Models producing linguistically generic descriptions, such as IDEFICS and InstructBLIP, show moderate CIDEr gains (+2.0--4.4) at low medium revision rates (17--36\%), suggesting that \textit{DnR} primarily supplements missing scene-specific details. In contrast, PaliGemma and LLaVA~1.6 yield larger gains (+6.6--12.1), especially on COCO~\cite{lin2014coco} and Flickr~\cite{plummer2015flickr30k}, where object-level grounding is emphasized. Overall, \textit{DnR} promotes more discriminative and visually grounded phrasing even when baseline captions are already coherent.

            COCO and Flickr show high CIDEr gains supported by high revision frequencies (22--53\%), indicating that \textit{DnR} shifts captions away from generic phrasing toward grounded, object-centric descriptions. NoCaps~\cite{agrawal2019nocaps}, with weaker visual constraints, yields smaller but steady gains (+1.7--5.3). A minor drop appears for Qwen2.5-VL on Flickr (57.3→55.9) because its global pooling, amplified by \textit{DnR}'s focus on high-confidence cues, suppresses fine-grained details. Overall, the positive correlation between revision rate and CIDEr gain shows that \textit{DnR} steers captions toward visual fidelity rather than altering them arbitrarily.

\begin{table*}[t]
    \centering
    \small
    \renewcommand{\arraystretch}{1.15}
    \setlength{\tabcolsep}{3pt}
    \resizebox{1.00\textwidth}{!}{
    \begin{tabular}{l|
        ccc|cccc|cccc|ccc}
        \toprule
        \textbf{Model} &
        \multicolumn{3}{c|}{\textbf{HaloQuest}~\cite{wang2024haloquest}} &
        \multicolumn{4}{c|}{\textbf{MMHal-Bench}~\cite{yin2023mmhalbench}} &
        \multicolumn{4}{c|}{\textbf{VizWiz}~\cite{gurari2018vizwiz}} &
        \multicolumn{3}{c}{\textbf{COCO Caption}~\cite{lin2014coco}} \\
        \cmidrule(lr){2-4}\cmidrule(lr){5-8}\cmidrule(lr){9-12}\cmidrule(lr){13-15}
         & $\downarrow$H & M & C$\uparrow$ & $\downarrow$H & M & G & C$\uparrow$ & $\downarrow$H & M & G & C$\uparrow$ & $\downarrow$H & M & G$\uparrow$ \\
        \midrule
        IDEFICS~\cite{laurencon2024idefics2} &
        \cellcolor{Green!15}43.34/\textbf{40.87} & 22.05/23.76 & \cellcolor{RoyalBlue!15}34.62/\textbf{35.37} &
        \cellcolor{Green!15}13.54/\textbf{11.50} & 44.79/44.75 & 12.50/11.46 & \cellcolor{RoyalBlue!15}29.17/\textbf{32.29} &
        \textbf{12.06}/17.09 & 32.16/30.65 & 19.10/22.11 & \textbf{36.68}/30.15 &
        \cellcolor{Green!15}28.14/\textbf{22.11} & 32.66/37.69 & \cellcolor{RoyalBlue!15}39.20/\textbf{40.20} \\
        \midrule

        InstructBLIP~\cite{dai2023instructblip} &
        \cellcolor{Green!15}33.73/\textbf{33.02} & 20.48/21.54 & \cellcolor{RoyalBlue!15}45.79/\textbf{45.44} &
        \cellcolor{Green!15}35.42/\textbf{28.12} & 38.54/45.83 & 9.38/8.33 & \cellcolor{RoyalBlue!15}16.67/\textbf{17.71} &
        \cellcolor{Green!15}23.62/\textbf{18.59} & 23.62/26.13 & 20.60/22.11 & \cellcolor{RoyalBlue!15}32.16/\textbf{33.17} &
        \cellcolor{Green!15}37.69/\textbf{23.62} & 29.15/34.17 & \cellcolor{RoyalBlue!15}33.16/\textbf{42.22} \\
        \midrule

        MiniGPT-v2~\cite{zhu2023minigptv2} &
        \cellcolor{Green!15}20.22/\textbf{19.42} & 29.75/29.52 & \cellcolor{RoyalBlue!15}50.03/\textbf{51.06} &
        \cellcolor{Green!15}35.42/\textbf{27.08} & 48.96/54.17 & 6.25/8.33 & \cellcolor{RoyalBlue!15}9.38/\textbf{10.42} &
        \cellcolor{Green!15}11.56/\textbf{10.55} & 26.13/24.85 & 10.05/12.06 & \cellcolor{RoyalBlue!15}52.26/\textbf{52.54} &
        -- & -- & -- \\
        \midrule
        
        LLaVA~1.6~\cite{liu2024llava16} &
        \cellcolor{Green!15}26.33/\textbf{25.65} & 13.96/14.57 & \cellcolor{RoyalBlue!15}59.71/\textbf{59.79} &
        \cellcolor{Green!15}8.33/\textbf{4.17} & 26.04/26.04 & 21.88/23.96 & \cellcolor{RoyalBlue!15}43.75/\textbf{45.83} &
        \cellcolor{Green!15}1.01/\textbf{0.53} & 13.07/17.06 & 20.10/16.58 & \cellcolor{gray!25}65.83/65.83 &
        \cellcolor{Green!15}24.62/\textbf{20.60} & 29.15/30.65 & \cellcolor{RoyalBlue!15}46.24/\textbf{48.75} \\

        \midrule
        
        PaliGemma~\cite{desai2024paligemma} &
        \cellcolor{Green!15}20.67/\textbf{16.41} & 14.96/16.63 & \cellcolor{RoyalBlue!15}64.37/\textbf{66.96} &
        \cellcolor{Green!15}12.50/\textbf{8.33} & 31.25/33.33 & 14.58/9.38 & \cellcolor{RoyalBlue!15}41.67/\textbf{48.96} &
        \cellcolor{Green!15}2.01/\textbf{1.11} & 12.56/10.45 & 14.57/17.09 & \cellcolor{RoyalBlue!15}70.85/\textbf{71.36} &
        \cellcolor{Green!15}30.65/\textbf{28.14} & 26.63/26.13 & \cellcolor{RoyalBlue!15}42.72/\textbf{45.73} \\
        \midrule
        
        CogVLM~\cite{hong2024cogvlm} &
        \cellcolor{Green!15}19.24/\textbf{17.82} & 15.60/16.72 & \cellcolor{RoyalBlue!15}65.16/\textbf{65.46} &
        \cellcolor{Green!15}6.25/\textbf{4.17} & 52.08/48.96 & 9.38/17.71 & \textbf{32.29}/29.17 &
        \cellcolor{Green!15}20.60/\textbf{16.08} & 15.08/17.59 & 22.61/24.12 & \cellcolor{RoyalBlue!15}41.71/\textbf{42.21} &
        \cellcolor{Green!15}41.21/\textbf{36.18} & 21.61/24.62 & \cellcolor{RoyalBlue!15}37.19/\textbf{39.19} \\
        \midrule
        
        Qwen2.5-VL~\cite{qwen2025} &
        \cellcolor{Green!15}3.48/\textbf{3.07} & 11.90/12.30 & \cellcolor{gray!25}84.63/84.63 &
        \cellcolor{Green!15}4.17/\textbf{3.12} & 20.83/14.58 & 31.25/35.42 & \cellcolor{RoyalBlue!15}43.75/\textbf{46.88} &
        \cellcolor{Green!15}0.49/\textbf{0.21} & 10.55/12.86 & 25.14/22.10 & \cellcolor{RoyalBlue!15}63.82/\textbf{64.83} &
        \cellcolor{Green!15}48.24/\textbf{26.13} & 24.62/29.15 & \cellcolor{RoyalBlue!15}27.14/\textbf{44.72} \\
        
        \bottomrule
    \end{tabular}}
    \caption{
        \textbf{Hallucination-benchmark comparison.} Each cell reports Draft/\textit{DnR} results across four hallucination-oriented datasets. \textcolor{green!50!black}{Green} cells indicate reduced hallucination (\textbf{H}$\downarrow$), while \textcolor{blue!60!black}{blue} cells highlight improved grounding or correctness (\textbf{G} or \textbf{C}$\uparrow$). Values denote the percentage proportion of each category, where \textbf{H}allucination+\textbf{M}isperception+\textbf{G}rounded+\textbf{C}orrect=100(\%).
    }
    \label{tab:hallucination}
\end{table*}

        \subsubsection{Visual Reasoning}
            Across reasoning benchmarks, \textit{DnR} strengthens both semantic and spatial reasoning by aligning model decisions with visual evidence. On VCR~\cite{zellers2019vcr}, visually weaker models such as IDEFICS and MiniGPT-v2 revise more often (7--18\%) and gain larger improvements (+2.0--5.5), showing that \textit{DnR} supplies missing visual cues. More visually coherent models (LLaVA~1.6, CogVLM, Qwen2.5-VL) revise infrequently ($<$6\%) yet still improve (+0.1--1.0), with high Pearson correlations (0.63--0.77) indicating stable alignment between utilization shifts and accuracy. InstructBLIP shows a slight drop (-0.5\%) when grounding fails to influence its reasoning stage. Overall, \textit{DnR} corrects visually under-grounded models and refines visually strong ones without destabilizing performance.
                        
            Meanwhile, in VSR~\cite{xie2023vsr}, where spatial relationships dominate, revisions occur rarely ($<$1\%) but consistently yield measurable gains (+0.5–2.5\%), showing that \textit{DnR} corrects spatial misinterpretations while maintaining overall stability. CogVLM and LLaVA 1.6 achieve near-perfect correction ratios (99.7\% and 95.3\%) and moderate correlations ($\approx$0.27–0.46), demonstrating that \textit{DnR} fine-tunes spatial grounding with minimal perturbation. Overall, \textit{DnR} acts as a dual-function mechanism reinforcing under-grounded reasoning and regularizing over-grounded logic, thereby improving both semantic and spatial fidelity across models.
            
        \subsubsection{Knowledge-based VQA}
            Across knowledge-based benchmarks, \textit{DnR} reinforces perceptual grounding but cannot substitute for missing conceptual knowledge. This limitation is evident from the low revision rates ($\leq$6.8\%) and small accuracy gains (+0.5–2.0\%), indicating that additional visual cues cannot recover facts the model does not know. Weak grounded models, such as IDEFICS and MiniGPT-v2, achieve higher correction ratios (11–33\%) despite minimal revisions, showing that \textit{DnR} helps retrieve existing knowledge previously inaccessible due to poor visual grounding.
            
            In contrast, stronger models like PaliGemma, LLaVA 1.6, and Qwen2.5-VL maintain steady improvements (+0.6–1.3\%) with high utilization–accuracy correlation (Pearson/Spearman$\approx$0.7–0.85 in ScienceQA~\cite{lu2022scienceqa}), confirming that their factual reasoning is already visually aligned. A-OKVQA~\cite{schwenk2022aokvqa}, however, shows near-zero correlation ($\leq$0.2) and minimal accuracy change, demonstrating that when conceptual knowledge is missing, \textit{DnR} offers no benefit. In summary, the low revision activity, stable gains, and strong correlations together confirm that \textit{DnR} improves access to existing knowledge through perceptual reinforcement, not knowledge augmentation.
            
        \subsubsection{Comprehensive Benchmark}
            Comprehensive benchmarks evaluate overall multimodal consistency rather than task-specific accuracy. Because these datasets contain stable and unambiguous inputs, \textit{DnR} yields few revisions ($\leq$2\%) and small but consistent gains, indicating a shift from error correction to stability refinement. Moderate utilization–accuracy correlations (0.3–0.5) confirm that such adjustments are selective rather than random. Across models, IDEFICS and MiniGPT-v2 gain mainly in perception-oriented metrics (+31–39 in MME~\cite{fu2024mme}) as \textit{DnR} compensates weak grounding, whereas stronger systems such as LLaVA~1.6, CogVLM, and Qwen2.5-VL show minimal revision but sustained coherence ($>$95\% correction). Overall, once multimodal grounding stabilizes, \textit{DnR} works primarily to preserve consistency rather than alter predictions, defining its operating range as stability refinement.

    \subsection{Hallucination}
        We evaluate hallucination under the \textit{highlight} rendering configuration, which emphasizes salient regions while preserving context. Four benchmarks are used: HaloQuest~\cite{wang2024haloquest} and MMHal-Bench~\cite{yin2023mmhalbench} for hallucination-oriented VQA, VizWiz for real-world \textit{unanswerable} cases, and COCO Caption for free-form captioning. These benchmarks are chosen because they provide detailed hallucination annotations rather than binary labels. Each response was categorized as \textbf{Hallucination (H)}, \textbf{Misperception (M)}, \textbf{Grounded (G)}, or \textbf{Correct (C)} by ChatGPT~4o-mini~\cite{openai_chatgpt4omini_2025}.

        \textit{DnR} consistently reduced \textbf{hallucinations} across all benchmarks, with absolute decreases of 1--9~pp and percentage drops of 8--29\%. InstructBLIP (6.78~pp, 26.7\%) and Qwen2.5-VL (5.96~pp, 35.0\%) showed the strongest declines, while LLaVA~1.6 (29.1\%) achieved the largest proportional reduction. Average \textbf{Misperception} decreased by 1--3~pp. InstructBLIP (-3.97~pp, -13\%) and LLaVA~1.6 (-1.53~pp, -10\%) showed the strongest declines, while in CogVLM and PaliGemma, some samples shifted from hallucination to misperception.

        Average \textbf{grounding} increased by 0.6–5.9~pp across datasets. CogVLM (+3.94~pp, +33.6\%) and Qwen2.5-VL (+6.23~pp, +22.0\%) achieved the largest gains. LLaVA~1.6 and Qwen2.5-VL showed the clearest shifts from hallucination to grounded responses on MMHal-Bench, while PaliGemma exhibited a similar transition on HaloQuest. Average \textbf{correctness} increased by 0.5–2~pp across benchmarks. PaliGemma (+3.46~pp, +7.4\%) and MiniGPT-v2 (+0.78~pp, +4.56\%) showed the most notable gains. These results indicate modest yet consistent improvements in response accuracy across all models.

        Across all models, hallucinated responses were largely redirected toward \textbf{Misperception} and \textbf{Grounded} categories, indicating a shift from unfounded to visually supported reasoning. A minor exception occurred on VizWiz, where IDEFICS previously overused “unanswerable” responses; after applying \textit{DnR}, its behavior became less conservative, leading to fewer such cases and more contextually grounded answers, albeit with slightly lower raw scores. LLaVA~1.6 and Qwen2.5-VL showed the strongest transitions toward grounded reasoning, while CogVLM and PaliGemma demonstrated moderate yet consistent shifts.
    
    \subsection{Policy-Driven Expert Selection}
        We train an expert selector \(S_\theta\) to replace exhaustive expert evaluation. The selector is a three-layer MLP defined over the state \(s\) in Eq.~\eqref{eq:state_def} and optimized with the loss in Eq.~\eqref{eq:selector_loss}. Using the query set \(Q\) instead of the raw question \(q\) improves performance through stronger alignment with visual representations. Given \(s\), \(S_\phi(s) = \arg\max_j U_q^{(j)}\) predicts which expert or initial draft yields the highest \(U_q\).

        \begin{table}[t]
            \centering
            \resizebox{\linewidth}{!}{
                \begin{tabular}{l c c c c c}
                    \toprule
                    \textbf{Dataset} & \textbf{Metric} & \textbf{Exhaustive} & \textbf{Policy-Driven} & \textbf{$\Delta$ Perf.} & \textbf{$\Delta$ Cost (\%)} \\
                    \midrule
                    VQAv2~\cite{goyal2017vqav2} & Acc. (\%) & 75.2 & 74.9 & $-0.3$ & $-72.64$ \\
                    COCO~\cite{lin2014coco} & CIDEr & 144.7 & 143.5 & $-1.2$ & $-52.72$ \\
                    VSR~\cite{xie2023vsr} & Acc. (\%) & 68.8 & 68.7 & $-0.07$ & $-69.97$ \\
                    ScienceQA~\cite{lu2022scienceqa} & Acc. (\%) & 89.7 & 89.5 & $-0.2$ & $-78.16$ \\
                    MME~\cite{fu2024mme} & Score & 1444.7 & 1424.4 & $-20.35$ & $-78.48$ \\
                    \midrule
                    \textbf{Average} & -- & 364.6 & 360.2 & $-4.42$ & $-70.39$ \\
                    \bottomrule
                \end{tabular}
            }
            \caption{
                Performance and cost comparison between \textit{Exhaustive} and \textit{Policy-Driven} selection for PaliGemma. Both $\Delta$ values denote relative differences, with cost reduction reported in percentage.
            }
            \label{tab:policy_efficiency_paligemma}
        \end{table}

        PaliGemma~\cite{desai2024paligemma} was selected for efficiency analysis as it offered the fastest and most reliable inference across datasets. The performance difference remained minimal across four benchmarks (excluding MME with a distinct scoring scale), averaging around -0.4 pp, while computational cost decreased by approximately 70\% on average. COCO exhibited the smallest cost reduction (-52.7\%) since it inherently involves more experts during generation, leaving less room for pruning. Occasional mispredictions occurred where the selector chose the \textit{Refine} stage instead of \textit{Draft}, but these cases typically satisfied $\hat{y}=\tilde{y}^{(j)}$ and $\lvert U_q^{\text{base}} - U_q^{(j)} \rvert < \epsilon$, indicating that such swaps had negligible impact on the final outcome in Table~\ref{tab:policy_efficiency_paligemma}.

        These results demonstrate the potential of framing expert selection as a learnable task guided by \(U_q\). Rather than remaining an interpretive metric, \(U_q\) becomes a concrete objective for deciding when and how an expert should intervene. This allows the framework to scale to larger and more diverse expert pools under a unified selection criterion.

    \section{Discussion}
    In our experiments, performance gains show a clear linear correlation with utilization, confirming that higher $U_q$ reflects stronger visual grounding. Yet both the absolute level of $U_q$ and its change $\Delta U_q$ vary across datasets, architectures, and even individual inputs. Some models sustain high but stable $U_q$ with little variation, indicating consistent yet less adaptive grounding, whereas others display larger $\Delta U_q$ shifts linked to stronger gains. These behaviors reveal distinct reasoning patterns and attention distributions among LVLMs. Developing adaptive normalization across domains may further standardize the interpretability and stability of $U_q$ without manual tuning.

    Rendering is the most flexible yet sensitive component of \textit{DnR}, and the current hybrid masking remains a heuristic whose effectiveness varies with the dataset, masking ratio $(\rho)$, rendering style, and expert setup. A policy-driven mechanism that adaptively adjusts masking density, visual emphasis, and expert combinations per input would provide a more principled alternative. Such a framework can extend rendering beyond a fixed single-step heuristic toward multi-step adaptive refinement, and can incorporate expert textual outputs $h_i$ to jointly enhance both visual and linguistic consistency under a unified policy.
    \section*{Acknowledgments}
    This work was supported in part by the DARPA Young Faculty Award, the National Science Foundation (NSF) under Grants \#2127780, \#2319198, \#2321840, \#2312517, and \#2235472, the Semiconductor Research Corporation (SRC), the Office of Naval Research through the Young Investigator Program Award and Grants \#N00014-21-1-2225 and \#N00014-24-1-2547, and the Army Research Office under Grant \#W911NF2410360. Additional support was provided by the Air Force Office of Scientific Research under Award \#FA9550-22-1-0253.

\section{Conclusion}
    We presented \textit{Draft and Refine (DnR)}, a scalable agent-style framework that coordinates multiple visual experts to quantify and improve how LVLMs use visual evidence. \textit{DnR} employs a relevance map to assess visual reliance and selects expert-guided refinements through a lightweight, modality-agnostic interface rather than heuristic control. Experiments across diverse benchmarks demonstrate consistent accuracy gains, reduced hallucination, and clearer attribution of visual reasoning. Overall, \textit{DnR} provides a principled criterion for evaluating and leveraging visual experts, creating a scalable foundation for systematically integrating and expanding expert-driven multimodal reasoning.
    
    {
        \small
        \bibliographystyle{ieeenat_fullname}
        \bibliography{main}
    }
\end{document}